\documentclass{sip}

\usepackage{epstopdf, epsfig} 
\usepackage{multirow}
\usepackage{graphicx}
\usepackage{amsmath}
\usepackage{amsxtra}
\usepackage{threeparttable}
\usepackage{amsmath}
\usepackage{amssymb}
\usepackage{subfig}

\usepackage{algorithm}
\usepackage{algorithmic}

\begin{document}

\title{Learning Priors for Adversarial Autoencoders}

\author[Hui-Po Wang, \textit{et al}.]{Hui-Po Wang$^{1}$, Wen-Hsiao Peng$^{1}$ and Wei-Jan Ko$^{1}$}

\address{\add{1}{National Chiao Tung University, 1001 Ta-Hsueh Rd., Hsinchu 30010, Taiwan}}

\corres{\name{Wen-Hsiao Peng}
\email{wpeng@cs.nctu.edu.tw}}

\begin{abstract}
Most deep latent factor models choose simple priors for simplicity, tractability
or not knowing what prior to use. Recent studies show that the choice of
the prior may have a profound effect on the expressiveness of the model,
especially when its generative network has limited capacity. In this paper, we propose to learn a proper prior from data for adversarial autoencoders
(AAEs). We introduce the notion of code generators to transform manually selected
simple priors into ones that can better characterize the data distribution. Experimental results show that the proposed model can generate better image quality and learn better disentangled representations than
AAEs in both supervised and unsupervised settings. Lastly, we present its
ability to do cross-domain translation in a  text-to-image synthesis task.
\end{abstract}

\keywords{Authors should not add keywords, as these will be chosen during the submission process (please refer to Section II (F) below for further details).}

\maketitle

\section{Introduction}
\label{sec:intro}
Deep latent factor models, such as variational autoencoders (VAEs) and adversarial autoencoders (AAEs), are becoming increasingly popular in various tasks, such as image generation \citep{larsen2015autoencoding}, unsupervised clustering \citep{dilokthanakul2016deep,makhzani2015adversarial}, and cross-domain translation \citep{wu2016learning}. These models involve specifying a prior distribution over latent variables and defining a deep generative network (i.e. the decoder) that maps latent variables to data space in stochastic or deterministic fashion. Training such deep models usually requires learning a recognition network (i.e. the encoder) regularized by the prior.

Traditionally, a simple prior, such as the standard normal distribution \citep{kingma2013auto}, is used for tractability, simplicity, or not knowing what prior to use. It is hoped that this simple prior will be transformed somewhere in the deep generative network into a form suitable for characterizing the data distribution. While this might hold true when the generative network has enough capacity, applying the standard normal prior often results in over-regularized models with only few active latent dimensions \citep{burda2015importance}. 

Some recent works \citep{hoffman2016elbo,goyal2017nonparametric,tomczak2017vae} suggest that the choice of the prior may have a profound impact on the expressiveness of the model. As an example, in learning the VAE with a simple encoder and decoder, Hoffman and Johnson \cite{hoffman2016elbo}
conjecture that multimodal priors can achieve a higher variational lower bound on the data log-likelihood than is possible with the standard normal prior. Tomczak and Welling \cite{tomczak2017vae} confirm the truth of this conjecture by showing that their multimodal prior, a mixture of the variational posteriors, consistently outperforms simple priors 
on a number of datasets in terms of maximizing the data log-likelihood. Taking one step further, Goyal~\etal \cite{goyal2017nonparametric} learn a tree-structured nonparametric Bayesian prior for capturing the hierarchy of semantics presented in the data. All these priors are learned under the VAE framework following the principle of maximum likelihood.  
      
Along a similar line of thinking, we propose in this paper the notion of
code generators for learning a prior from data for AAE. The objective is to learn a code generator network to transform a simple prior into one that,  together with the generative network, can better characterize the data distribution. To this end, we generalize the framework of AAE in several significant ways: 
\begin{itemize}
        \item We replace the simple prior with a learned prior by training the code generator to output latent variables that will minimize an adversarial loss in data space.  

        \item We employ a learned similarity metric \citep{larsen2015autoencoding} in place of the default squared error in data space for training the autoencoder.    
        \item We maximize the mutual information between part of the code generator input and the decoder output  for supervised and unsupervised training using a variational technique introduced in InfoGAN \citep{chen2016infogan}.  
\end{itemize}

Extensive experiments confirm its effectiveness of generating better quality images and learning better disentangled representations than AAE in both supervised and unsupervised settings, particularly on complicated datasets. In addition, to the best of our knowledge, this is one of the first few works that attempt to introduce a learned prior for AAE.

The remainder of this paper is organized as follows: Section~\ref{sec:related_work} reviews the background and related works. Section~\ref{sec:method} presents the implementation details and the training procedure of the proposed code generator. Section~\ref{sec:experiments} presents extensive experiments to show the superiority of our models over prior works. Section~\ref{sec:text2img} showcases an application of our model to text-to-image synthesis. Lastly, we conclude this paper with remarks on future work. 

\section{Related Work}
\label{sec:related_work}
\begin{figure}[tbp]
\centering
\includegraphics[width=0.45\textwidth]{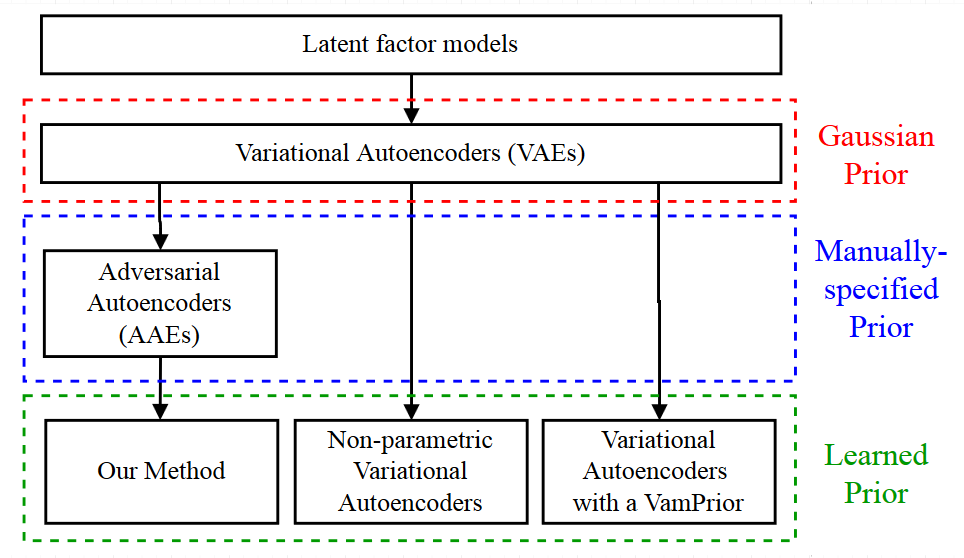}
\caption{The relations of our work with prior arts.}
\label{fig:relations}
\end{figure}
A latent factor model is a probabilistic model for describing the relationship between a set of latent and visible variables. The model is usually specified by a prior distribution $p(z)$ over the latent variables $z$ and a conditional distribution $p(x|z;\theta)$ of the visible variables $x$ given the latent variables $z$. The model parameters $\theta$ are often learned by maximizing the marginal log-likelihood of the data $\log p(x;\theta)$. 

\textbf{Variational Autoencoders (VAEs).} To improve
the model's expressiveness, it is common to make deep the conventional latent
factor model by introducing a neural network
to $p(x|z;\theta)$. One celebrated example is VAE \citep{kingma2013auto}, which assumes the following prior $p(z)$ and $p(x|z;{\theta})$:
        \begin{equation}
        \begin{split}
        \label{eq:vae_x_z}
          p(z) & \sim \mathcal{N}(z;0,I) \\
          p(x|z;{\theta}) & \sim \mathcal{N}(x;o(z;\theta),\sigma^{2}I)
        \end{split}
        \end{equation}
where the mean $o(z;\theta)$ is modeled by the output of a neural network
with parameters $\theta$. In this case, the marginal $p(x;\theta)$
becomes intractable; the model is thus trained by maximizing the log
evidence lower-bound (ELBO): 
        \begin{equation}
        \label{eq:vae_loss}
        \mathcal{L}(\phi, \theta) = E_{q(z|x;\phi)} \log p(x|z;\theta) - KL(q(z|x;\phi) \parallel p(z))
        \end{equation}
where $q(z|x;\phi)$ is the variational density, implemented by another neural
network with parameter $\phi$, to approximate the posterior $p(z|x;\theta)$. When regarding $q(z|x;\phi)$  as an (stochastic)\ encoder and $p(z|x;\theta)$
 as a (stochastic)\ decoder, Equation (\ref{eq:vae_loss}) bears an interpretation of training an autoencoder with the latent code $z$ regularized by the prior $p(z)$ through the KL-divergence.

\textbf{Adversarial Autoencoders (AAEs).} Motivated by the observation that VAE is largely limited by the Gaussian prior assumption, i.e. $p(z) \sim \mathcal{N}(z;0,I)$, Makhzani~\etal \cite{makhzani2015adversarial} relax this constraint by allowing $p(z)$ to be any distribution. Apparently, the
KL-divergence becomes intractable when $p(z)$ is arbitrary. They
thus replace the KL-divergence with an adversarial loss imposed on the
encoder output, requiring  that the latent code $z$ produced by the encoder should have an aggregated posterior distribution\footnote{The aggregated posterior distribution is defined as $q(z) = \int q(z|x;\phi)p_{d}(x)dx$,
where $p_d(x)$ denotes the empirical distribution of the training data.} 
the same as the prior $p(z)$.    



\textbf{Non-parametric Variational Autoencoders (Non-parametric VAEs).} While
AAE allows the prior to be arbitrary, how to select a prior that can best
characterize the data distribution remains an open issue. Goyal~\etal \cite{goyal2017nonparametric} make an attempt to learn a non-parametric prior based on the nested Chinese restaurant process for VAEs. Learning is achieved by fitting it to the aggregated posterior distribution, which amounts to maximization of ELBO. The result induces a hierarchical structure of semantic concepts in latent space. 

\textbf{Variational Mixture of Posteriors (VampPrior).} The VampPrior \cite{tomczak2017vae} is
a new type of prior for the VAE. It consists of a mixture of the variational posteriors conditioned on a set of learned pseudo-inputs $\{x_k\}$. In symbol, this prior is given by
        \begin{equation}
        \label{eq:VampPrior}
            p(z)=\frac{1}{K}\sum_{k=1}^K q(z|x_k;\phi)
        \end{equation}   
Its multimodal nature and coupling with the posterior achieve superiority over many other simple priors in terms of training complexity and expressiveness.   

Inspired by these learned priors \citep{goyal2017nonparametric,tomczak2017vae} for VAE,  we propose in this paper the notion of code generators to learn a proper prior from data for AAE. The relations of our work with these prior arts are illustrated in Fig.~\ref{fig:relations}.  


\newlength\myindent
\setlength\myindent{2em}
\newcommand\bindent{%
  \begingroup
  \setlength{\itemindent}{\myindent}
  \addtolength{\algorithmicindent}{\myindent}
}
\newcommand\eindent{\endgroup}

\section{Method}
\label{sec:method}

\begin{figure}[tbp]
\setlength\tabcolsep{2pt}
    \centering
    \subfloat[]{%
    \includegraphics[width=\linewidth]{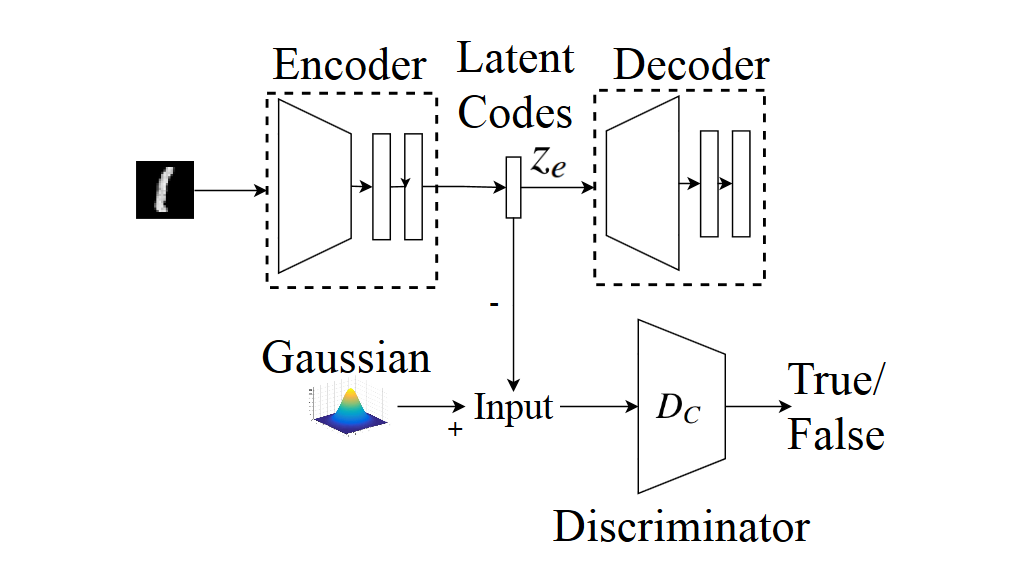}%
    }%
    \\
    \subfloat[]{%
    \includegraphics[width=\linewidth]{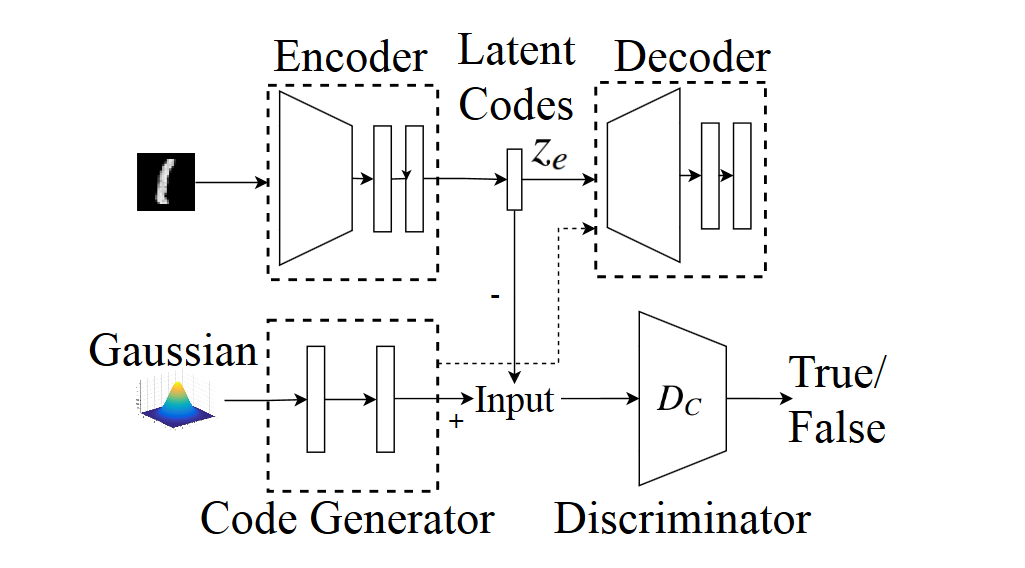}%
    }%
\caption{The architecture of AAE without (a) and with (b) the code generator.}
\label{fig:aae_and_cg}
\end{figure}

In this paper, we propose to learn the prior from data instead of specifying
it arbitrarily. Based on the framework of AAE, we introduce a
neural network (which we call the \textit{code generator}) to transform the manually-specified prior into a better form. Fig.~\ref{fig:aae_and_cg} presents its role in the overall architecture, and contrasts the architectural difference relative to AAE.

\subsection{Learning the Prior}
Because the code generator itself has to be learned, we need an objective function to shape the distribution at its output. Normally, we wish to find a prior that, together with the decoder (see Fig.~\ref{fig:aae_and_cg} (b)), would lead to a prior distribution that maximizes the data likelihood. We are however faced with two challenges. First, the output of the code generator could be any distribution, which may make the likelihood function and its variational lower bound intractable. Second, the decoder has to be learned simultaneously, which creates a moving target for the code generator.

\begin{figure}[tbp]
\setlength\tabcolsep{1pt}
    \centering
    \subfloat[]{%
    \includegraphics[width=\linewidth]{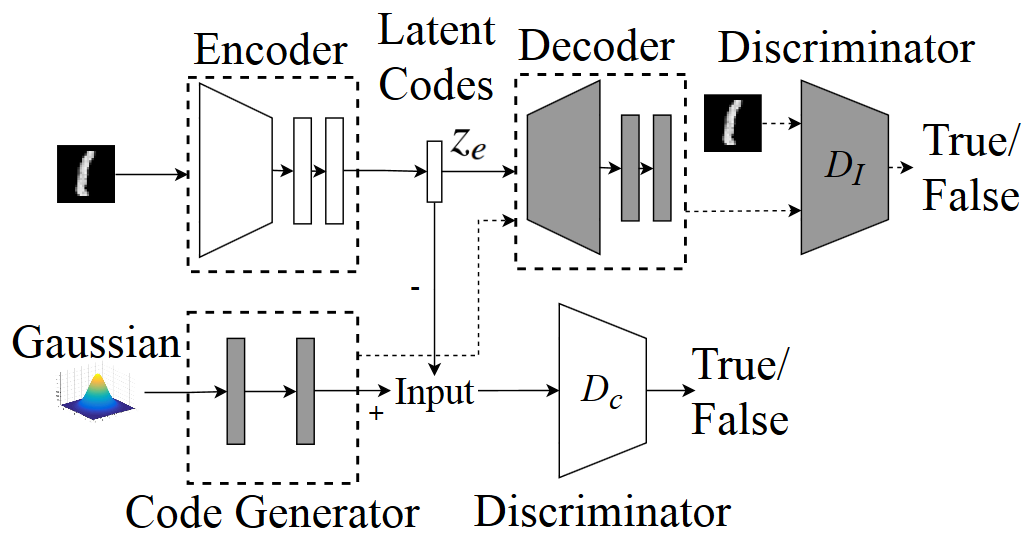}%
    }%
    \\
    \subfloat[]{%
    \includegraphics[width=\linewidth]{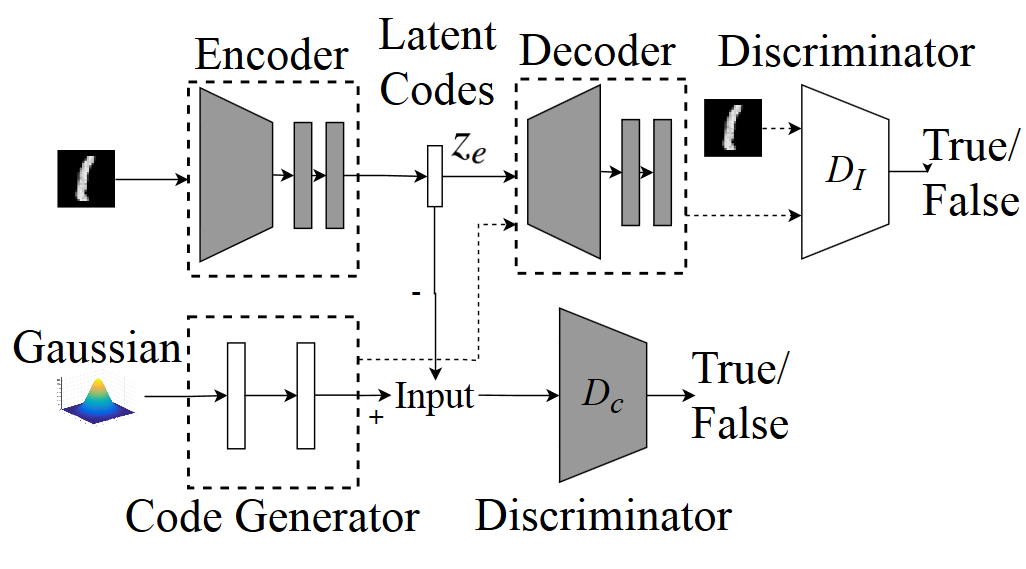}%
    }%
\caption{Alternation of training phases: (a) the prior improvement phase and (b) the AAE phase. The shaded building blocks indicate the blocks to be updated.}
\label{fig:model_phases}
\end{figure}

To address the first challenge, we propose to impose an adversarial loss on the output of the decoder when training the code generator. That is, we want the code generator to produce a prior distribution that minimizes the adversarial loss at the decoder output. Consider the example in Fig.~\ref{fig:model_phases} (a). The decoder should generate images with a distribution that in principle matches the empirical distribution of real images in the training data, when driven by the output of the code generator. In symbols, this is to minimize  
    \begin{equation}
    \label{eq:CGloss}
    \mathcal{L}^{I}_{GAN} = \log(D_{I}(x)) + \log(1-D_I(dec(z_c))),     
    \end{equation}
where $z_c=CG(z)$ is the output of the code generator $CG$ driven by a noise sample $z \sim p(z)$, $D_I$ is the discriminator in image
space, and $dec(z_c)$ is the output of the decoder driven by $z_c$.

To address the second challenge, we propose to alternate the training
of the code generator and the decoder/encoder until convergence. In one phase,
termed the \textit{prior improvement phase} (see Fig.~\ref{fig:model_phases} (a)), we update the code generator with the loss function in Eq.~(\ref{eq:CGloss}), by fixing the encoder\footnote{Supposedly,
the decoder needs to be fixed in this phase. It
is however found beneficial in terms of convergence to update also the decoder.}. In the other phase, termed the \textit{AAE phase} (see Fig.~\ref{fig:model_phases} (b)), we fix the code generator and update the autoencoder following the training procedure of AAE. Specifically, the encoder output has to be regularized by minimizing the following adversarial loss:   
        \begin{equation}
        \label{eq:dis_code_loss}
        \mathcal{L}^C_{GAN} = \log(D_{C}(z_c)) + \log(1-D_C(enc(x))),  
        \end{equation}
where $z_c=CG(z)$ is the output of the code generator, $enc(x)$ is the encoder output given the input $x$, and $D_C$ is the discriminator
in latent code space. 

Because the decoder will be updated in both phases, the convergence of the decoder relies on consistent training objectives of the two training phases. It is however noticed that the widely used pixel-wise squared error criterion in the AAE phase tends to produce blurry decoded images. This obviously conflicts with the adversarial objective in the prior improvement phase, which requires the decoder to produce sharp images. Inspired by the notion of learning similarity metrics \citep{larsen2015autoencoding} and perceptual loss \citep{johnson2016perceptual}, we change the criterion of minimizing squared error in pixel domain to be in feature domain. Specifically, in the AAE phase, we require that a reconstructed image $dec(enc(x))$ should minimize squared error in feature domain with respect to its original input $x$. This loss is referred to as \textit{perceptual loss} and is defined by
\begin{equation}
    \mathcal{L}_{rec} = \|\mathcal{F}(dec(enc(x)))-\mathcal{F}(x)\|^2, 
\end{equation}
where $\mathcal{F}(\cdot)$ denotes the feature representation (usually the output of the last convolutional layer in the image discriminator $D_I$) of an image. With this, the decoder would be driven consistently in both phases towards producing decoded images that resemble
closely real images.

\begin{figure}[tbp]
\begin{center}
\includegraphics[width=\linewidth]{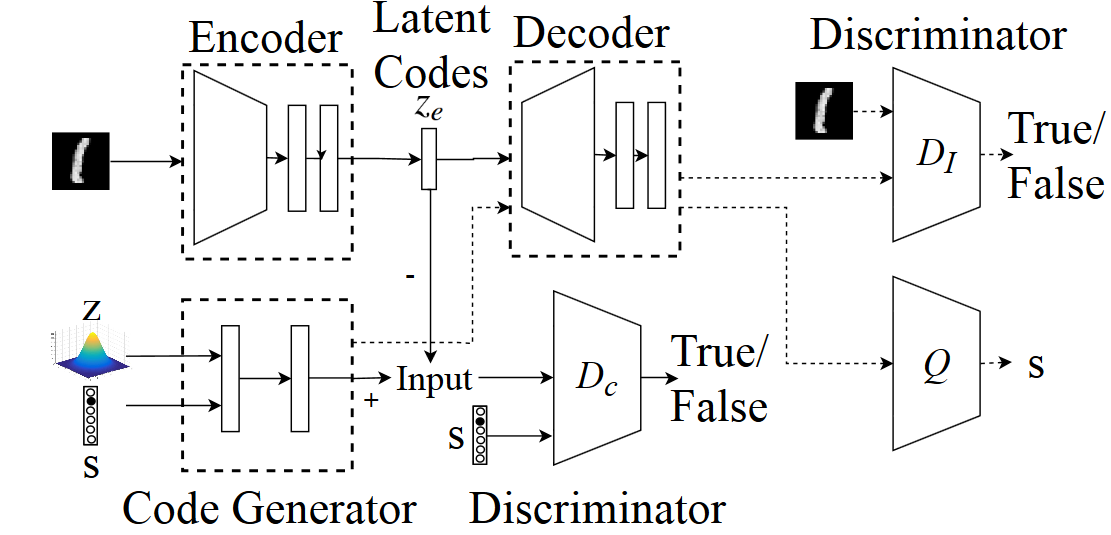}
\end{center}
\caption{Supervised learning architecture with the code generator.}
\label{fig:supervised_models}
\end{figure}

\begin{figure}[tbp]
\begin{center}
\includegraphics[width=\linewidth]{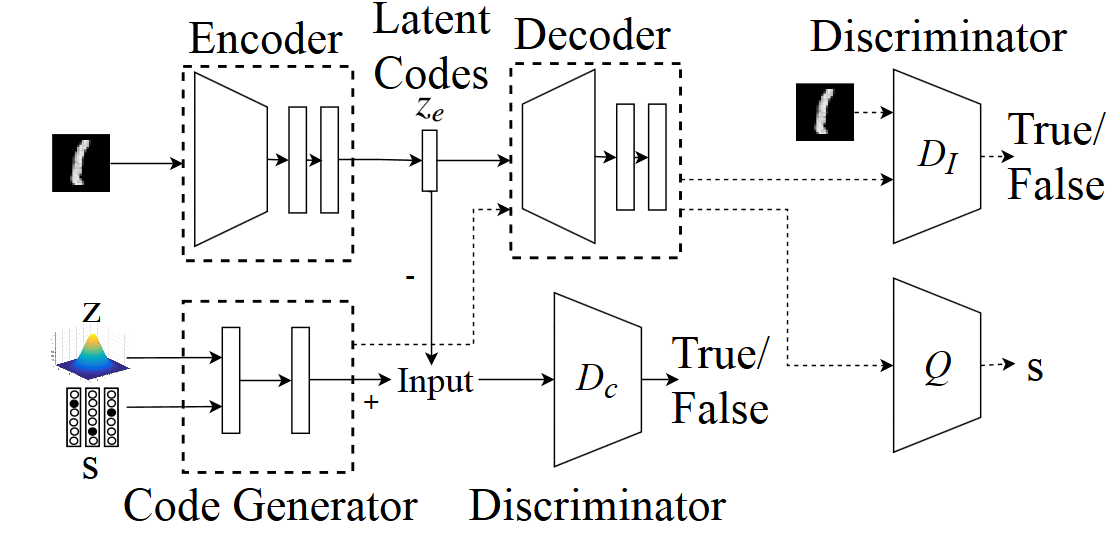}
\end{center}
\caption{Unsupervised learning architecture with the code generator.}
\label{fig:unsupervised_models}
\end{figure}

\subsection{Learning Conditional Priors}

\subsubsection{Supervised Setting}
The architecture in Fig.~\ref{fig:model_phases} can be extended to learn conditional priors supervisedly. Such priors find applications in conditional data generation, e.g. conditional image generation in which the decoder generates images according to their class labels $s$. To this end, we make three major changes to the initial architecture:
\begin{itemize}
    \item Firstly, the code generator now takes as inputs a data label $s$ and a noise variable $z$ accounting for the intra-class variety, and produces a prior distribution conditional on the label $s$ (see Fig.~\ref{fig:supervised_models}).
    
    \item Secondly, the end-to-end mutual information $I(s;dec(z_{c}))$ between the label $s$ and the decoded image $dec(z_{c})$ is maximized as part of our training objective to have both the code generator and the decoder pick up the information carried by the label variable $s$ when generating the latent code $z_{c}$ and subsequently the decoded image $dec(z_{c})$. This is achieved by maximizing its variational lower bound $\mathcal{L}_{I}(s; dec(z_{c}))$ of $I(s;dec(z_{c}))$ \citep{chen2016infogan} as given by 
    \begin{equation}
    \begin{split}
        \label{eq:lower_bound}    
        &\mathcal{L}_{I}(s; dec(z_{c})) \\
        &= -E_{z, s \sim p(z, s), z_{c} \sim CG(z, s)}[\log Q(s|dec(z_{c}))],
    \end{split}
    \end{equation}
where $p(z, s)=p(z)p(s)$ is the joint distribution of the label $s$ and the noise $z$, $CG(\cdot)$ is the code generator, and $Q(s|dec(z_{c}))$ is a classifier used to recover the label $s$ of the decoded image $dec(z_{c})$.

  \item Lastly, the discriminator $D_{c}$ in latent code space is additionally provided with the label $s$ as input, to implement class-dependent regularization at the encoder output during the AAE learning phase. That is, 
        \begin{equation}
        \label{eq:supervised_dis_code_loss}
        \mathcal{L}^C_{GAN} = \log(D_{C}(z_c, s)) + \log(1-D_C(enc(x), s)),   
        \end{equation}
where $s$ is the label associated with the input image $x$. 
\end{itemize}

The fact that the label $s$ of an input image $x$ needs to be properly fed to different parts of the network during training indicates the supervised learning nature of the aforementioned procedure. 



\subsubsection{Unsupervised Setting}
Taking one step further, we present in Fig.~\ref{fig:unsupervised_models} a re-purposed architecture to learn conditional priors under an unsupervised setting. Unlike the supervised setting where the correspondence between the label $s$ and the image $x$ is explicit during training, the unsupervised setting is to learn the correspondence in an implicit manner. Two slight changes are thus made to the architecture in Fig. \ref{fig:supervised_models}: (1) the label $s$ at the input of the code generator is replaced with a label drawn randomly from a categorical distribution; and (2) the discriminator $D_c$ in the latent code space is made class agnostic by removing the label input. The former is meant to produce a multimodal distribution in the latent space while the latter is to align such a distribution with that at the encoder output. Remarkably, which mode (or class) of distribution an image $x$ would be assigned to in latent code space is learned implicitly. In a sense, we hope the code generator can learn to discover the intriguing latent code structure inherent at the encoder output. It is worth pointing out that in the absence of any regularization or guidance, there is no guarantee that this learned assignment would be in line with the semantics attached artificially to each data sample. 

Algorithm~\ref{alg:our_training} details the training procedure.

        \begin{algorithm}[tbp]
        \caption{Training procedure.}
        \label{alg:our_training}
        \begin{algorithmic}
        \STATE {Initialize $\theta_{enc}$, $\theta_{dec}$, $\theta_{CG}$, $\theta_{D_I}$, $\theta_{D_C}$, $\theta_{Q}$}
        \STATE {Repeat (for each epoch $E_{i}$)}
        \STATE {\quad Repeat (for each mini-batch $x_{j}$)}
        \STATE {\qquad // AAE phase}
        \STATE {\qquad If label $s$ exists then}
        \bindent
        \STATE {\qquad $z, s \sim p(z, s)=p(z)p(s)$}
        \STATE {\qquad $z_c \leftarrow CG(z, s)$}
        \eindent
        \STATE {\qquad Else}
        \bindent
        \STATE {\qquad $z \sim p(z)$}
        \STATE {\qquad $z_c \leftarrow CG(z)$}
        \eindent
        \STATE {\qquad End If}
        \STATE {\qquad }
        \STATE {\qquad Compute $\mathcal{L}^{C}_{GAN}$, $\mathcal{L}_{rec}$}         
        \STATE {\qquad}
        \STATE {\qquad // Update network parameters}
        \STATE {\qquad $\theta_{D_C} \leftarrow \theta_{D_C} - \nabla_{\theta_{D_C}}(-\mathcal{L}^{C}_{GAN})$}
        \STATE {\qquad $\theta_{enc} \leftarrow \theta_{enc} - \nabla_{\theta_{enc}}( \mathcal{L}^{C}_{GAN} + \mathcal{L}_{rec})$}
        \STATE {\qquad $\theta_{dec} \leftarrow \theta_{dec} - \nabla_{\theta_{dec}}(\lambda * \mathcal{L}_{rec})$}
        \STATE {\qquad}
        \STATE {\qquad // Prior improvement phase}
        \STATE {\qquad If label $s$ exists then}
        \bindent
        \STATE {\qquad $z, s \sim p(z, s)=p(z)p(s)$}
        \STATE {\qquad $z_c \leftarrow CG(z, s)$}
        \STATE {\qquad Compute $\mathcal{L}^{I}_{GAN}$ and $\mathcal{L}_{I}(s;dec(z_{c}))$}
        \eindent
        \STATE {\qquad Else}
        \bindent
        \STATE {\qquad $z \sim p(z)$}
        \STATE {\qquad $z_c \leftarrow CG(z)$}
        \STATE {\qquad Compute $\mathcal{L}^{I}_{GAN}$}
        \eindent
        \STATE {\qquad End If}
        \STATE {\qquad}
        \STATE {\qquad // Update network parameters}
        \STATE {\qquad $\theta_{D_{I}} \leftarrow \theta_{D_{I}} - \nabla_{\theta_{D_{I}}}(-\mathcal{L}^{I}_{GAN})$}
        \STATE {\qquad If label $s$ exists then}
        \bindent
        \STATE {\qquad $\theta_{dec} \leftarrow \theta_{dec} - \nabla_{\theta_{dec}}(\mathcal{L}^{I}_{GAN} + \mathcal{L}_{I}(s;dec(z_{c}))$}
        \STATE {\qquad $\theta_{CG} \leftarrow \theta_{CG} - \nabla_{\theta_{CG}}(\mathcal{L}^{I}_{GAN} + \mathcal{L}_{I}(s;dec(z_{c}))$}
        \STATE {\qquad $\theta_{Q} \leftarrow \theta_{Q} - \nabla_{\theta_{Q}}(\mathcal{L}_{I}(s;dec(z_{c}))) $}
        
        \eindent
        \STATE {\qquad Else}
        \bindent
        \STATE {\qquad $\theta_{dec} \leftarrow \theta_{dec} - \nabla_{\theta_{dec}}(\mathcal{L}^{I}_{GAN})$}
        \STATE {\qquad $\theta_{CG} \leftarrow \theta_{CG} - \nabla_{\theta_{CG}}(\mathcal{L}^{I}_{GAN})$}
        \eindent
        \STATE {\qquad End If}
        \STATE {\quad Until all mini-batches are processed}
        \STATE {Until termination}
        \end{algorithmic}
        \end{algorithm}
\section{Experiments}
\label{sec:experiments}
We first show the superiority of our learned priors over manually-specified priors, followed by an ablation study of individual components. In the end, we compare the performance of our model with AAE in image generation tasks. Unless stated otherwise, all the models adopt the same autoencoder for a fair comparison.


\subsection{Comparison with Prior Works}

\begin{table}[tbp]
\centering
\caption{Comparison with AAE, VAE, and Vamprior on CIFAR-10}
\label{table:prior_choice}
\begin{tabular}{lc}
\hline
\textbf{Method} & \textbf{Inception Score} \\ \hline
AAE \cite{makhzani2015adversarial} w/ a Gaussian prior & 2.15 \\
VAE \cite{kingma2013auto} w/ a Gaussian prior & 3.00 \\
Vamprior \cite{tomczak2017vae} &  2.88\\ 
Our method w/ a learned prior & 6.52 \\ \hline
\end{tabular}
\end{table}

\begin{figure*}[tbp] 
    \centering
    \centering
    \subfloat[AAE \cite{makhzani2015adversarial}]{%
        \includegraphics[width=0.22\linewidth]{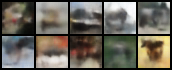}%
        }%
    \hspace{0.01\linewidth} 
    \centering
    \subfloat[VAE \cite{kingma2013auto}]{%
        \includegraphics[width=0.22\linewidth]{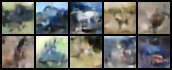}%
        }%
    \hspace{0.01\linewidth} 
    \centering
    \subfloat[Vamprior \cite{tomczak2017vae}]{%
        \includegraphics[width=0.22\linewidth]{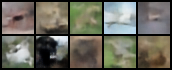}%
        }
    \hspace{0.01\linewidth} 
    \centering
    \subfloat[Proposed model]{%
        \includegraphics[width=0.22\linewidth]{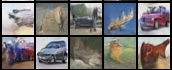}%
    }     
    \caption{Sample images produced by (a) AAE, (b) VAE, (c) Vamprior, and (d) the proposed model.}
    \label{fig:comparison_image}
\end{figure*}

Latent factor models with their priors learned from data rather than specified manually should better characterize the data distribution. To validate this, we compare the performance of our model with several prior arts, including AAE \citep{makhzani2015adversarial}, VAE \citep{kingma2013auto}, and Vamprior \citep{tomczak2017vae}, in terms of Inception Score (IS). Of these works, AAE chooses a Gaussian prior and regularizes the latent code distribution with an adversarial loss \cite{goodfellow2014generative}. VAE \cite{kingma2013auto} likewise adopts a Gaussian prior yet uses the KL-divergence for regularization. Vamprior \cite{tomczak2017vae} learns for VAE a Gaussian mixture prior. 
For the results of Vamprior \cite{tomczak2017vae}, we run their released software \cite{tomczak2017vae} but replace their autoencoder with ours for a fair comparison.

Table \ref{table:prior_choice} compares their Inception Score for image generation on CIFAR-10 with a latent code size of 64. As expected, both AAE \cite{makhzani2015adversarial} and VAE \cite{kingma2013auto}, which adopt manually-specified priors, have a lower IS of 2.15 and 3.00, respectively. Somewhat surprisingly, Vamprior \cite{tomczak2017vae}, although using a learned prior, does not have an advantage over VAE \cite{kingma2013auto} with a simple Gaussian prior in the present case. This may be attributed to the fact that the prior is limited to be a Gaussian mixture distribution. Relaxing this constraint by modeling the prior with a neural network, our model achieves the highest IS of 6.52.

Fig. \ref{fig:comparison_image} further visualizes sample images generated with these models by driving the decoder with latent codes drawn from the prior or the code generator in our case. It is observed that our model produces much sharper images than the others. This confirms that a learned and flexible prior is beneficial to the characterization and generation of data. 

\begin{figure*}[tbp] 
    \centering
    \centering
    \subfloat[BEGAN \cite{berthelot2017began}]{%
        \includegraphics[width=0.18\linewidth]{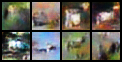}%
        }%
    \hspace{0.01\linewidth} 
    \centering
    \subfloat[DCGAN \cite{radford2015unsupervised}]{%
        \includegraphics[width=0.18\linewidth]{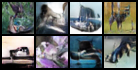}%
        }%
    \hspace{0.01\linewidth} 
    \centering
    \subfloat[LSGAN \cite{mao2017least}]{%
        \includegraphics[width=0.18\linewidth]{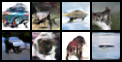}%
        }
    \hspace{0.01\linewidth} 
    \centering
    \subfloat[WGAN-GP \cite{gulrajani2017improved}]{%
        \includegraphics[width=0.18\linewidth]{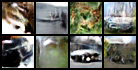}%
    }     
    \hspace{0.01\linewidth} 
    \centering
    \subfloat[Proposed model]{%
        \includegraphics[width=0.18\linewidth]{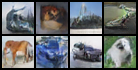}%
    }     
    \caption{Subjective quality evaluation of generated images produced by state-of-the-art generative models.}
    \label{fig:comparison_gan_image}
\end{figure*}

To get a sense of how our model performs as compared to other state-of-the-art generative models, Table \ref{is_table} compares their Inception Score on CIFAR-10. Caution must be exercised in interpreting these numbers as these models adopt different decoders (or generative networks). With the current implementation, our model achieves a comparable score to other generative models. Few sample images of these models are provided in Fig. \ref{fig:comparison_gan_image} for subjective evaluation.

\begin{table}[btp]
\centering
\caption{Comparison with other state-of-the-art generative models on CIFAR-10}
\label{is_table}
\begin{tabular}{lc}
\textbf{Method} & \textbf{Inception Score} \\ \hline
BEGAN \cite{berthelot2017began} & 5.62  \\
DCGAN  \cite{radford2015unsupervised} & 6.16 \\
LSGAN \cite{mao2017least} & 5.98 \\
WGAN-GP \cite{gulrajani2017improved} & 7.86 \\ 
Our method w/ a learned prior  & 6.52 \\ \hline
\end{tabular}
\end{table}

\subsection{Ablation Study}
\begin{table*}[tbp]
\caption{Inception score of generated images with the models trained on CIFAR-10: A, B, and C denote respectively the design choices of enabling the learned prior, the perceptual loss, and the updating of the decoder in both phases.}
\label{table:ablation_study}
\centering
\begin{tabular}{llllc}
\hline
Method & A & B & C & IS \\ \hline
AAE \cite{makhzani2015adversarial} w/ a Gaussian prior and MSE loss &  &  &  & 2.15 \\
AAE w/ a learned prior and MSE loss & \checkmark &  &  & 3.04 \\
AAE w/ a learned prior and perceptual loss & \checkmark & \checkmark &  & 3.69 \\
Ours (full) & \checkmark & \checkmark & \checkmark & 6.52 \\ \hline
 &  &  &  & 
\end{tabular}
\end{table*}

In this section, we conduct an ablation study to understand the effect of (A) the learned prior, (B) the perceptual loss, and (C) the updating of the decoder in both phases on Inception Score \cite{salimans2016improved}. To this end, we train an AAE \cite{makhzani2015adversarial} model with a 64-D Gaussian prior on CIFAR-10 as the baseline. We then enable incrementally each of these design choices. For a fair comparison, all the models are equipped with an identical autoencoder architecture yet trained with their respective objectives.

From Table \ref{table:ablation_study}, we see that the baseline has the lowest IS and that replacing the manually-specified prior with our learned prior increases IS by about 0.9. Furthermore, minimizing the perceptual loss instead of the conventional mean squared error in training the autoencoder achieves an even higher IS of 3.69. This suggests that the perceptual loss does help make more consistent the training objectives for the decoder in the AAE and the prior improvement phases. Under such circumstances, allowing the decoder to be updated in both phases tops the IS.

\begin{figure}[tbp] 
    \centering
    \centering
    \subfloat[Our model + 8-D latent code]{%
        \includegraphics[width=0.45\linewidth]{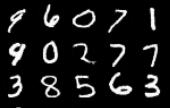}%
        }%
    \hspace{0.01\linewidth} 
    \centering
    \subfloat[AAE \cite{makhzani2015adversarial} + 8-D latent code]{%
        \includegraphics[width=0.45\linewidth]{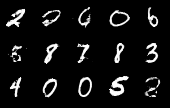}%
        }%
    \\    
    \centering
    \subfloat[Our model + 64-D latent code]{%
        \includegraphics[width=0.45\linewidth]{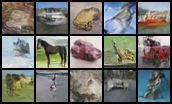}%
        }
    \hspace{0.01\linewidth} 
    \centering
    \subfloat[AAE \cite{makhzani2015adversarial} + 64-D latent code]{%
        \includegraphics[width=0.45\linewidth]{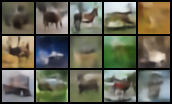}%
    }     
    \caption{Images generated by our model and AAE trained on MNIST (upper) and CIFAR-10 (lower).}
    \label{fig:image_sampling}
\end{figure}

\begin{figure}[tbp] 
    \centering
    \subfloat[Our model + 100-D latent code]{%
        \includegraphics[width=0.45\linewidth]{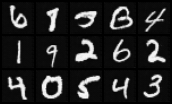}%
        }
    \hspace{0.01\linewidth} 
    \centering
    \subfloat[AAE \cite{makhzani2015adversarial} + 100-D latent code]{%
        \includegraphics[width=0.45\linewidth]{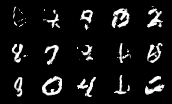}%
        }
        \\
    \centering
    \subfloat[Our model + 2000-D latent code]{%
        \includegraphics[width=0.45\linewidth]{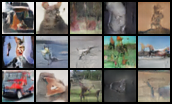}%
        }%
    \hspace{0.01\linewidth} 
    \centering
    \subfloat[AAE \cite{makhzani2015adversarial} + 2000-D latent code]{%
        \includegraphics[width=0.45\linewidth]{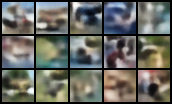}%
        }%
    \caption{Images generated by our model and AAE trained on MNIST (upper) and CIFAR-10 (lower). In this experiment, the latent code dimension is increased significantly to 64-D and 2000-D for MNIST and CIFAR-10, respectively. For AAE, the re-parameterization
trick is applied to the output of the encoder as suggested in \citep{makhzani2015adversarial}.}
    \label{fig:image_sampling_hd}
\end{figure}

\subsection{In-depth Comparison with AAE}
\label{ssec:image_generation}
Since our model is inspired by AAE \citep{makhzani2015adversarial}, this section provides an in-depth comparison with it in terms of image generation. In this experiment, the autoencoder in our model is trained based on minimizing the perceptual loss (i.e. the mean squared error in feature domain), whereas by convention, AAE \citep{makhzani2015adversarial} is trained by minimizing the mean squared error in data domain.

Fig.~\ref{fig:image_sampling} displays side-by-side images generated from these models when trained on MNIST and CIFAR-10 datasets. They are produced by the decoder driven by samples from their respective priors. In this experiment, two observations are immediate. First, our model can generate sharper images than AAE \citep{makhzani2015adversarial} on both datasets. Second, AAE \citep{makhzani2015adversarial} experiences problems in reconstructing visually-plausible images on the more complicated CIFAR-10. These highlight the advantages of optimizing the autoencoder with the perceptual loss and learning the code generator through an adversarial loss, which in general produces subjectively sharper images.

Another advantage of our model is its ability to have better adaptability in higher dimensional latent code space. Fig.~\ref{fig:image_sampling_hd} presents images generated by the two models when the dimension of the latent code is increased significantly from 8 to 100 on MNIST, and from 64 to 2000 on CIFAR-10. As compared to Fig.~\ref{fig:image_sampling}, it is seen that the increase in code dimension has little impact on our model, but exerts a strong influence on AAE \citep{makhzani2015adversarial}. In the present case, AAE \citep{makhzani2015adversarial} can hardly produce recognizable images, particularly on CIFAR-10, even after the re-parameterization trick has been applied to the output of the encoder as suggested in \citep{makhzani2015adversarial}. This emphasizes the importance of having a prior that can adapt automatically to a change in the dimensionality of the code space and data.

\subsection{Disentangled Representations}
\label{ssec:disentangle_rep}

Learning disentangled representations is desirable in many applications. It refers generally to learning a data representation whose individual dimensions can capture independent factors of variation in the data. To demonstrate the ability of our model to learn disentangled representations and the merits of data-driven priors, we repeat the disentanglement tasks in \citep{makhzani2015adversarial}, and compare its performance with AAE \citep{makhzani2015adversarial}.

\subsubsection{Supervised Learning}
\label{sssec:supervised_exp}

This section presents experimental results of using the network architecture in Fig.~\ref{fig:supervised_models} to learn supervisedly a code generator $CG$ that outputs a conditional prior given the image label $s$ for characterizing the image distribution. In particular, the remaining uncertainty about the image's appearance given its label is modeled by transforming a Gaussian noise $z$ through the code generator $CG$. By having the noise $z$ be independent of the label $s$, we arrive at a disentangled representation of images. At test time, the generation of an image $x$ for a particular class is achieved by inputting the class label $s$ and a Gaussian noise $z$ to the code generator $CG(\cdot)$ and then passing the resulting code $z_c$ through the decoder $x=dec(z_c)$. 


To see the sole contribution from the learned prior, the training of the AAE baseline \citep{makhzani2015adversarial} also adopts the perceptual loss and the mutual information maximization; that is, the only difference to our model is the direct use of the label $s$ and the Gaussian noise $z$ as the conditional prior.

\begin{figure*}[tbp]
\begin{center}
        \subfloat[Our model]{
                \includegraphics[width=0.45\linewidth]{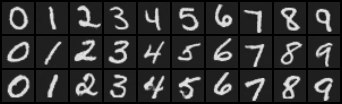}
        }
        \hspace{0.01\linewidth}
        \subfloat[AAE \cite{makhzani2015adversarial}]{
                \includegraphics[width=0.45\linewidth]{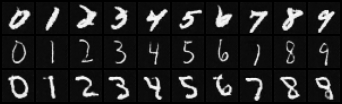}
        }
        \\
        \subfloat[Our model]{
                \includegraphics[width=0.45\linewidth]{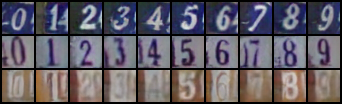}
        }
        \hspace{0.01\linewidth}
        \subfloat[AAE \cite{makhzani2015adversarial}]{
                \includegraphics[width=0.45\linewidth]{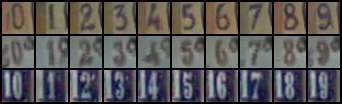}
        }
        \\
        \subfloat[Our model]{
                \includegraphics[width=0.45\linewidth]{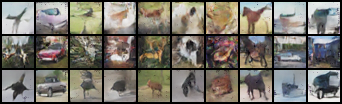}
        }
        \hspace{0.01\linewidth}
        \subfloat[AAE \cite{makhzani2015adversarial}]{
                \includegraphics[width=0.45\linewidth]{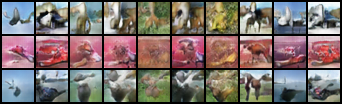}
        }
\end{center}
\caption{Images generated by the proposed model (a)(c)(e) and AAE (b)(d)(f) trained on MNIST, SVHN and CIFAR-10 datasets in the supervised setting. Each column of images have the same label/class information but varied Gaussian noise. On the other hand, each row of images  have the same Gaussian noise but varied label/class variables.}
\label{fig:supervised_disentangled_results}
\end{figure*}

Fig.~\ref{fig:supervised_disentangled_results}
displays images generated by our model and AAE \citep{makhzani2015adversarial}. Both models adopt a 10-D one-hot vector to specify the label $s$ and a 54-D Gaussian to generate the noise $z$. To be fair, the output of our code generator has an identical dimension (i.e. 64) to the latent prior of AAE \citep{makhzani2015adversarial}. Each row of Fig.~\ref{fig:supervised_disentangled_results} corresponds to images generated by varying the label $s$ while fixing the noise $z$. Likewise, each column shows images that share the same label $s$ yet with varied noise $z$.

On MNIST and SVHN, both models work well in separating the label information from the remaining (style) information. This is evidenced from the observation that along each row, the main digit changes with the label $s$ regardless of the noise $z$, and that along each column, the style varies without changing the main digit. On CIFAR-10, the two models behave differently. While both produce visually plausible images, ours generate more semantically discernible images that match their labels.

\begin{figure*}[tbp]
\begin{center}
        \subfloat[MNIST]{ 
                \includegraphics[width=0.33\linewidth]{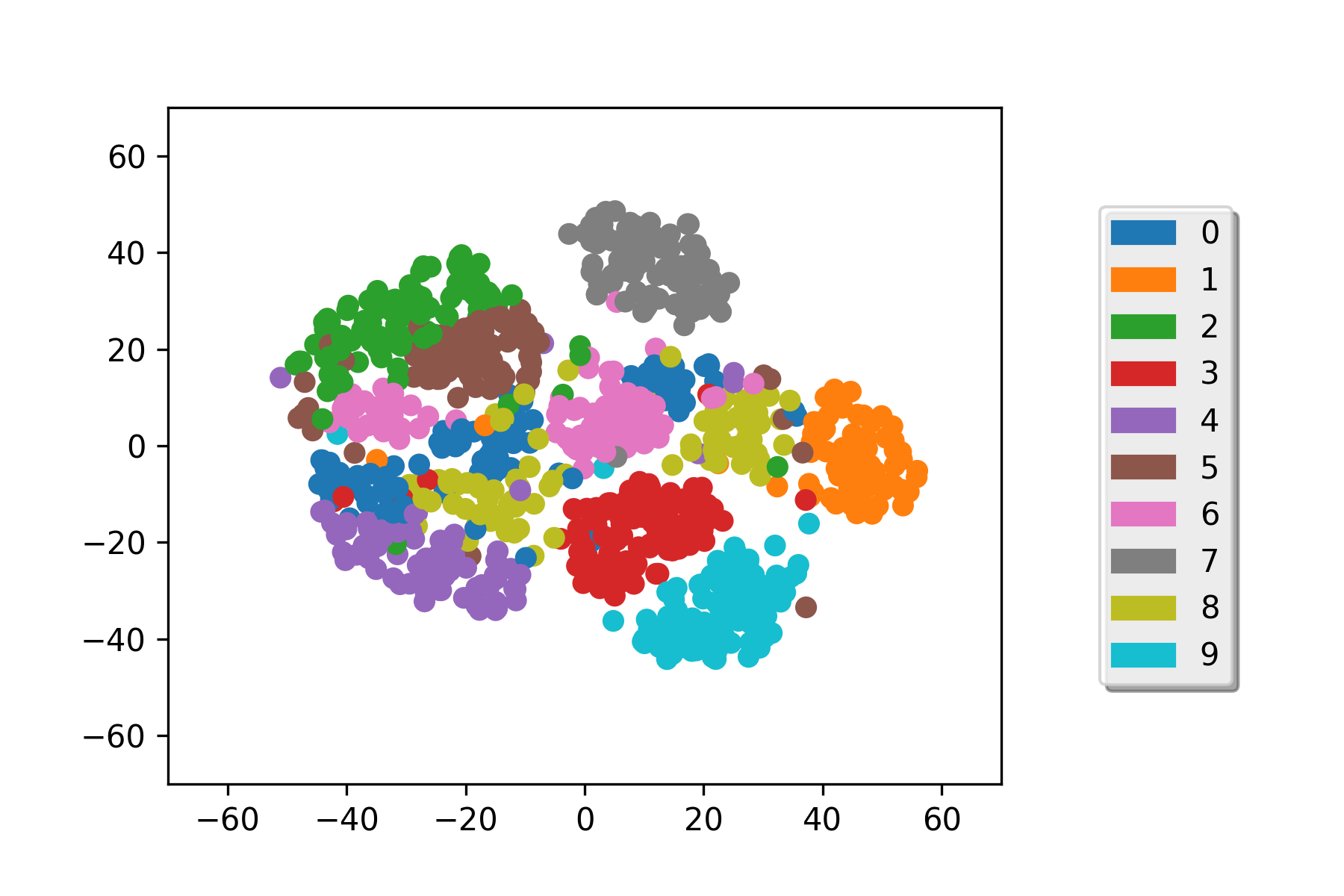}
        }
        \subfloat[SVHN]{
                \includegraphics[width=0.33\linewidth]{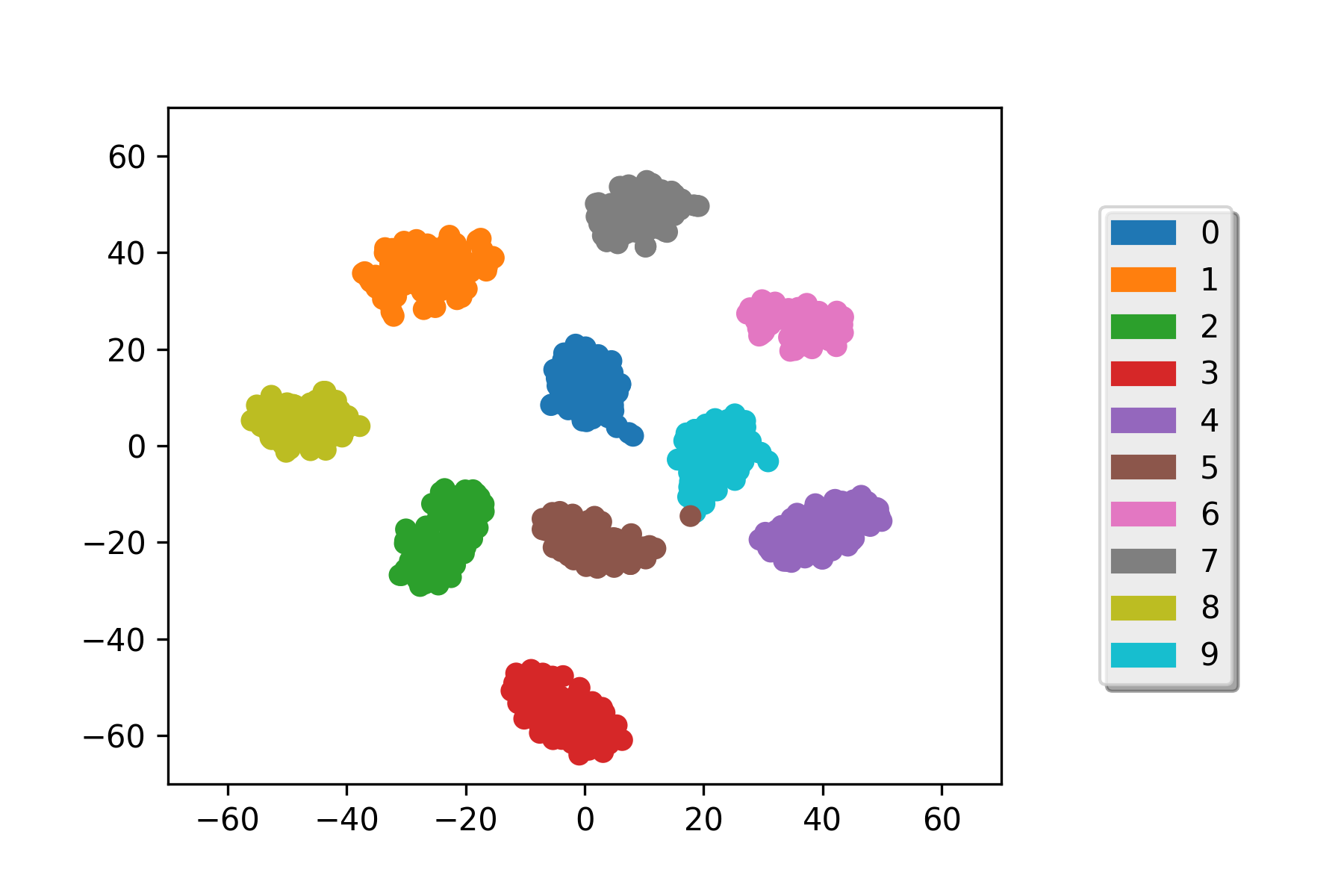}
        }
        \subfloat[CIFAR-10]{ 
                \includegraphics[width=0.33\linewidth]{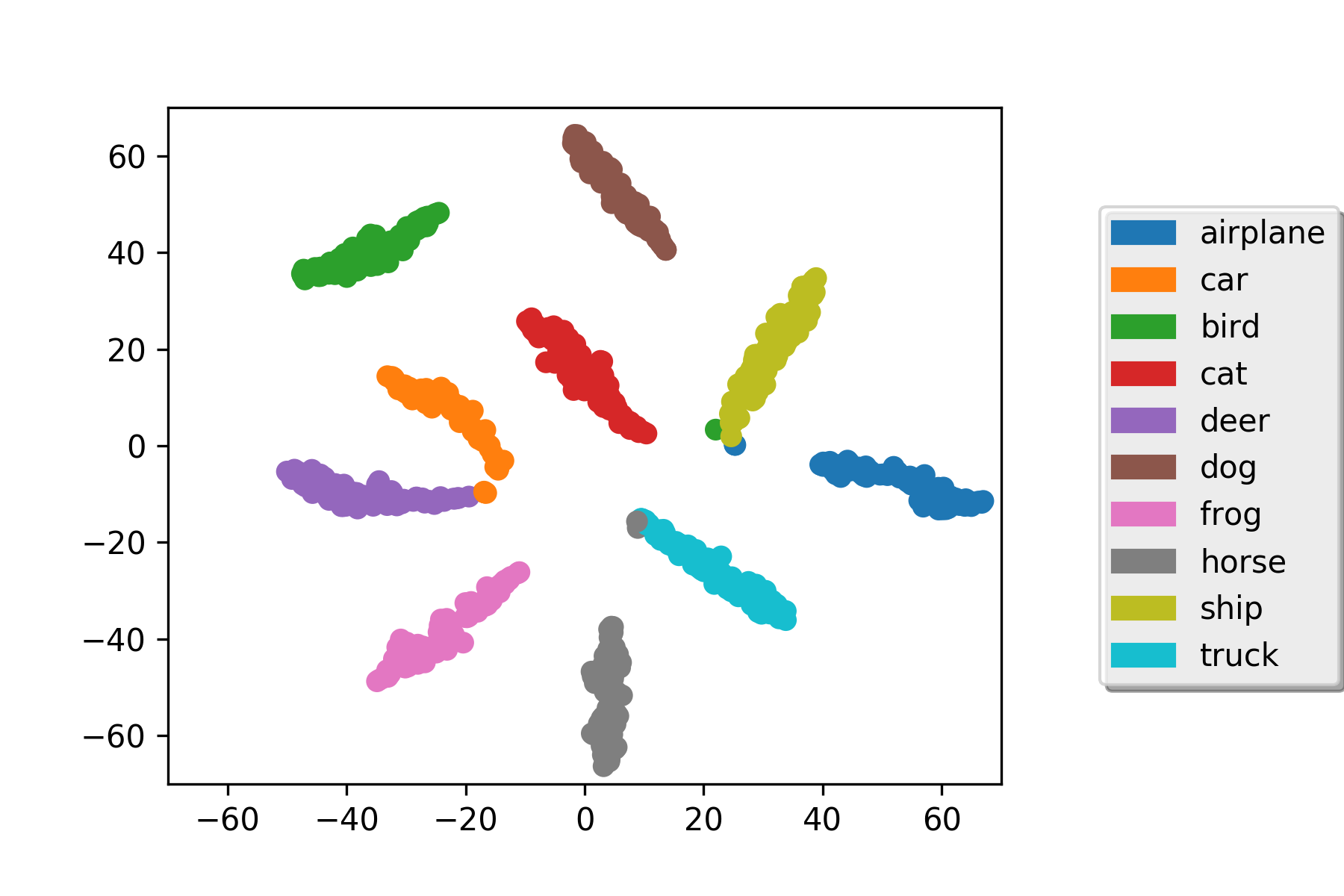}
        }
\caption{Visualization of the code generator output in the supervised setting.}
\label{fig:supervised_code_visualization}
\end{center}
\end{figure*}

Fig.~\ref{fig:supervised_code_visualization} visualizes the output of the code generator with the t-distributed stochastic neighbor embedding (t-SNE). It is seen that the code generator learns a distinct conditional distribution for each class of images. It is believed that the more apparent inter-class distinction reflects the more difficult it is for the decoder to generate images of different classes. Moreover, the elliptic shape of the intra-class distributions in CIFAR-10 may be ascribed to the higher intra-class variability.

\subsubsection{Unsupervised Learning}
\label{sssec:unsupervised_exp}

This section presents experimental results of using the network architecture in Fig.~\ref{fig:unsupervised_models} to learn unsupervisedly a code generator $CG$ that outputs a prior for best characterizing the image distribution. In the present case, the label $s$ is drawn randomly from a categorical distribution and independently from the Gaussian input $z$, as shown in Fig.~\ref{fig:unsupervised_models}. The categorical distribution encodes our prior belief about data clusters, with the number of distinct values over which it is defined specifying the presumed number of clusters in the data. The Gaussian serves to explain the data variability within each cluster. In regularizing the distribution at the encoder output, we want the code generator $CG$ to make sense of the two degrees of freedom for a disentangled representation of images.

At test time, image generation is done similarly to the supervised case. We start by sampling $s$ and $z$, followed by feeding them into the code generator and then onwards to the decoder. In this experiment, the categorical distribution is defined over 10-D one-hot vectors and the Gaussian is 90-D. As in the supervised setting, after the model is trained, we alter the label variable $s$ or the Gaussian noise $z$ one at a time to verify whether the model has learned to cluster images unsupervisedly. We expect that a good model should generate images with certain common properties (e.g. similar backgrounds or digit types) when the Gaussian part $z$ is altered while the label part $s$ remains fixed. 

\begin{figure*}[btp]
\begin{center}
        \subfloat[Our model]{ 
                \includegraphics[width=0.45\linewidth]{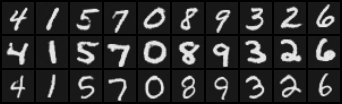}
        }
        \hspace{0.01\linewidth}
        \subfloat[AAE]{
                \includegraphics[width=0.45\linewidth]{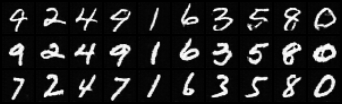}
        }
        \\
        \subfloat[Our model]{
                \includegraphics[width=0.45\linewidth]{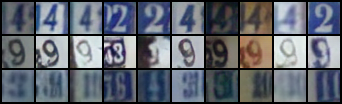}
        }
        \hspace{0.01\linewidth}
        \subfloat[AAE]{
                \includegraphics[width=0.45\linewidth]{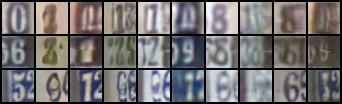}
        }
        \\
        \subfloat[Our model]{ 
                \label{fig:unsupervised_cifar10_our_method}
                \includegraphics[width=0.45\linewidth]{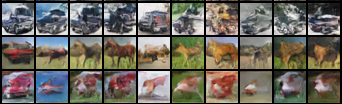}
        }
        \hspace{0.01\linewidth}
        \subfloat[AAE]{
                \label{fig:unsupervised_cifar10_AAE}
                \includegraphics[width=0.45\linewidth]{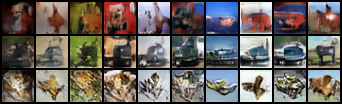}
        }
\end{center}
\caption{Images generated by the proposed model (a)(c)(e) and AAE (b)(d)(f) trained on MNIST, SVHN and CIFAR-10 datasets in the unsupervised setting. Each column of images have  the same label/class information but varied Gaussian noise. On the other hand, each row of images  have the same Gaussian noise but varied label/class variables.}
\label{fig:unsupervised_disentangled_results}
\end{figure*}

The results in Fig.~\ref{fig:unsupervised_disentangled_results} show that on MNIST, both our model and AAE successfully learn to disentangle the digit type from the remaining information. Based on the same presentation order as in the supervised setting, we see that each column of images (which correspond to the same label $s$) do show images of the same digit. On the more complicated SVHN and CIFAR-10 datasets, each column mixes images from different digits/classes. It is however worth noting that both models have a tendency to cluster images with similar backgrounds according to the label variable $s$. Recall that without any semantic guidance, there is no guarantee that the clustering would be in line with the semantics attached artificially to each data sample.

\begin{figure*}[tbp]
\begin{center}
        \subfloat[Encoder (MNIST)]{ 
                \includegraphics[width=0.33\linewidth]{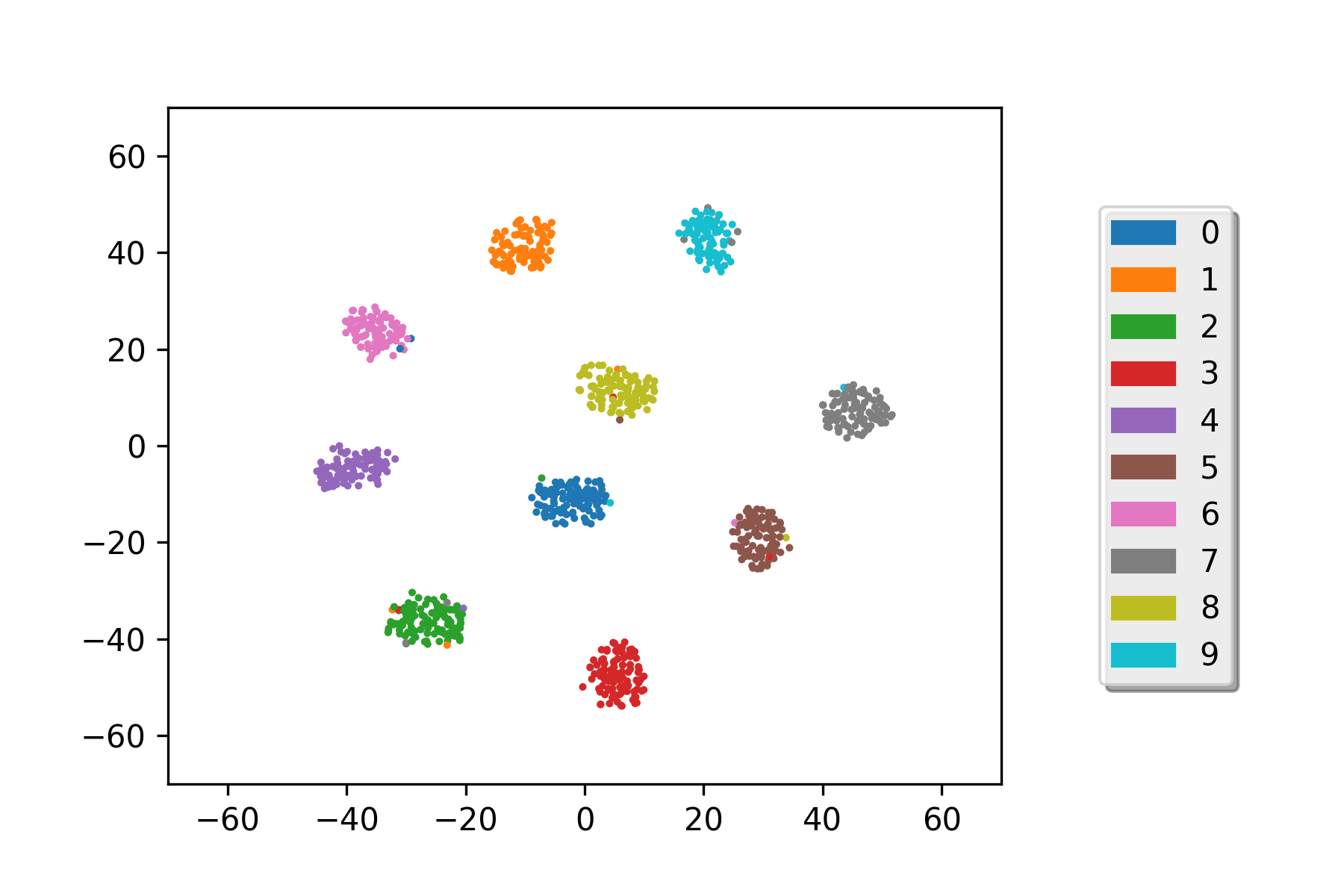}
        }
        \subfloat[Encoder (SVHN)]{
                \includegraphics[width=0.33\linewidth]{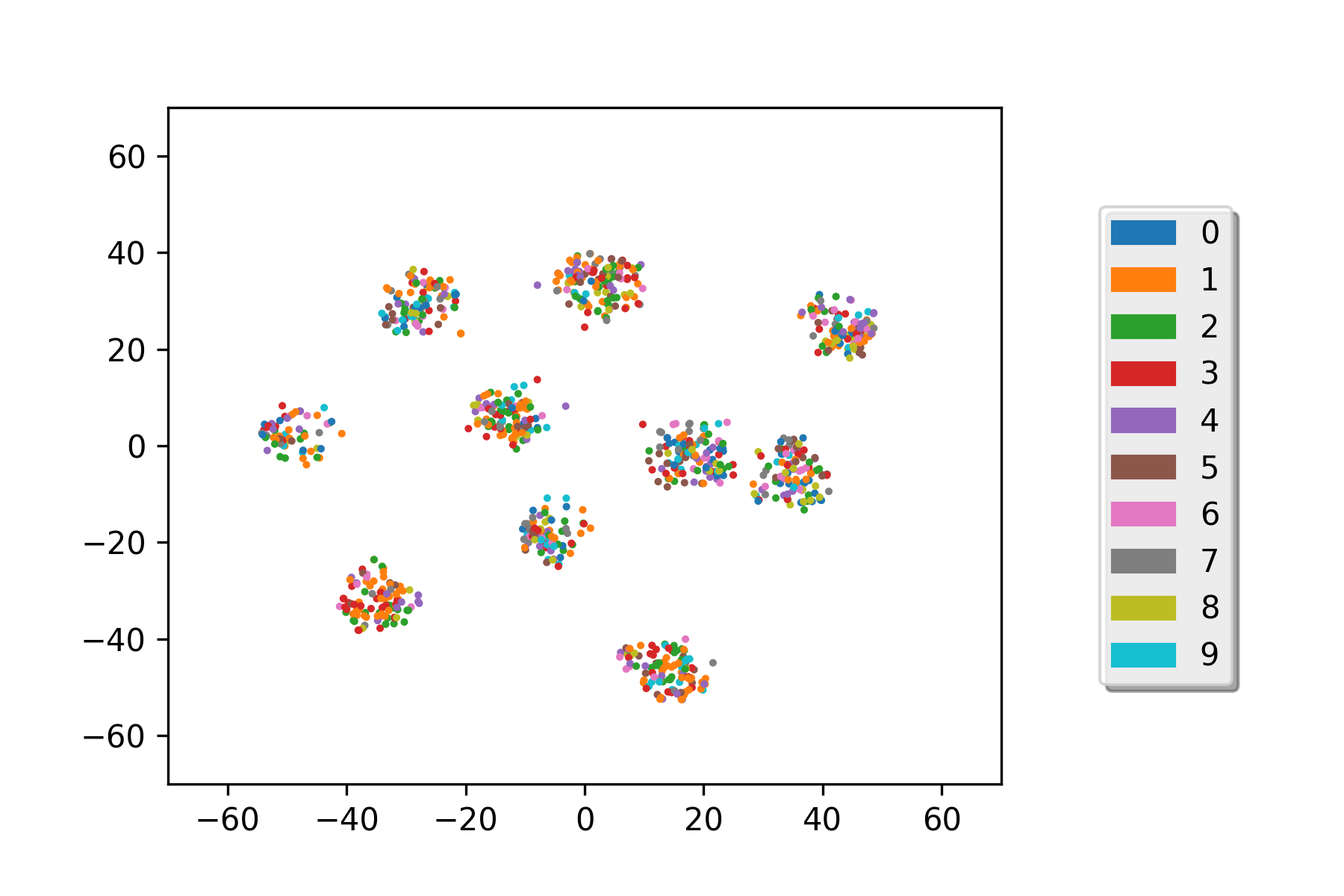}
        }
        \subfloat[Encoder (CIFAR-10)]{ 
                \includegraphics[width=0.33\linewidth]{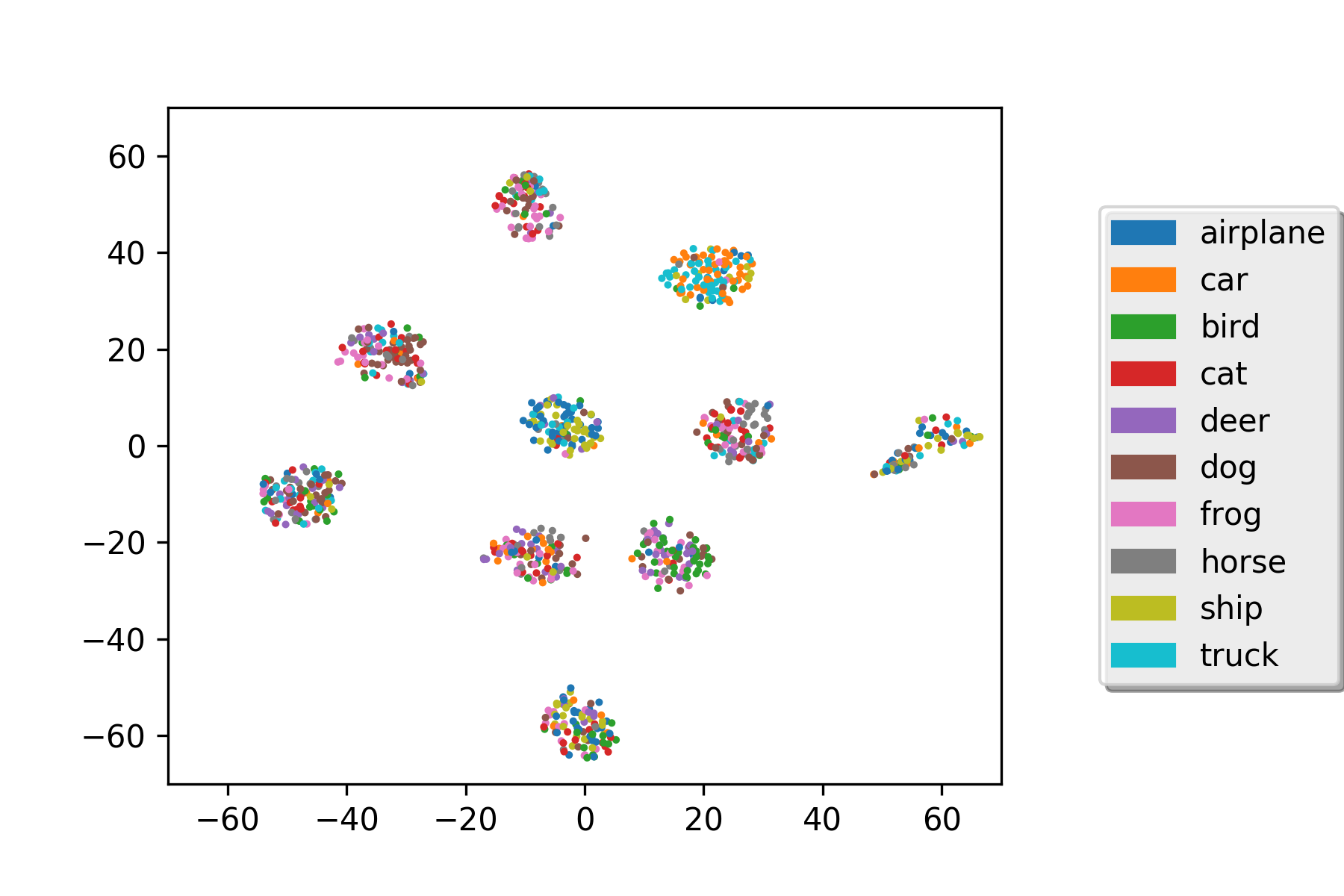}
        }
        \\
        \subfloat[Code generator (MNIST)]{ 
                \includegraphics[width=0.33\linewidth]{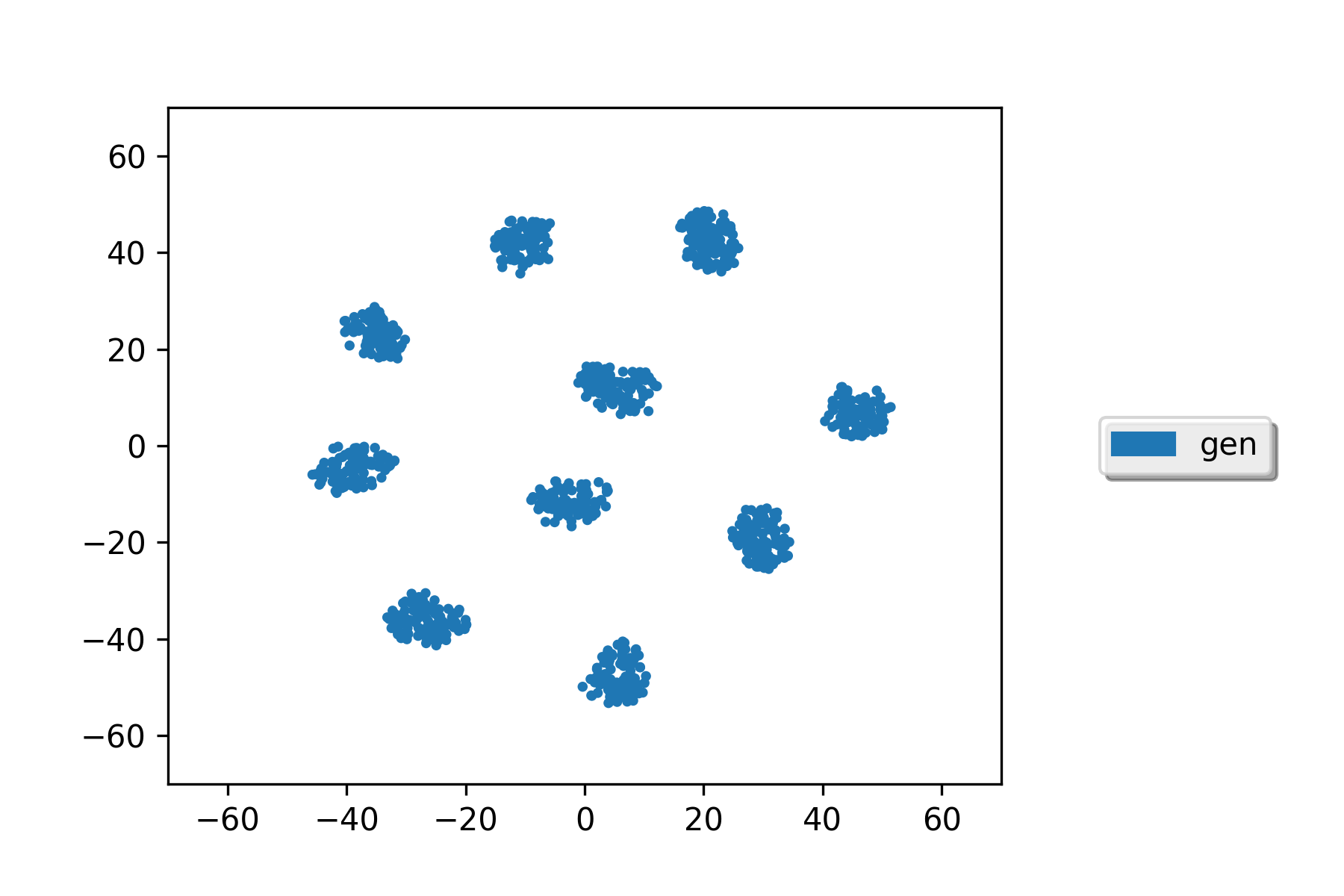}
        }
        \subfloat[Code generator (SVHN)]{
                \includegraphics[width=0.33\linewidth]{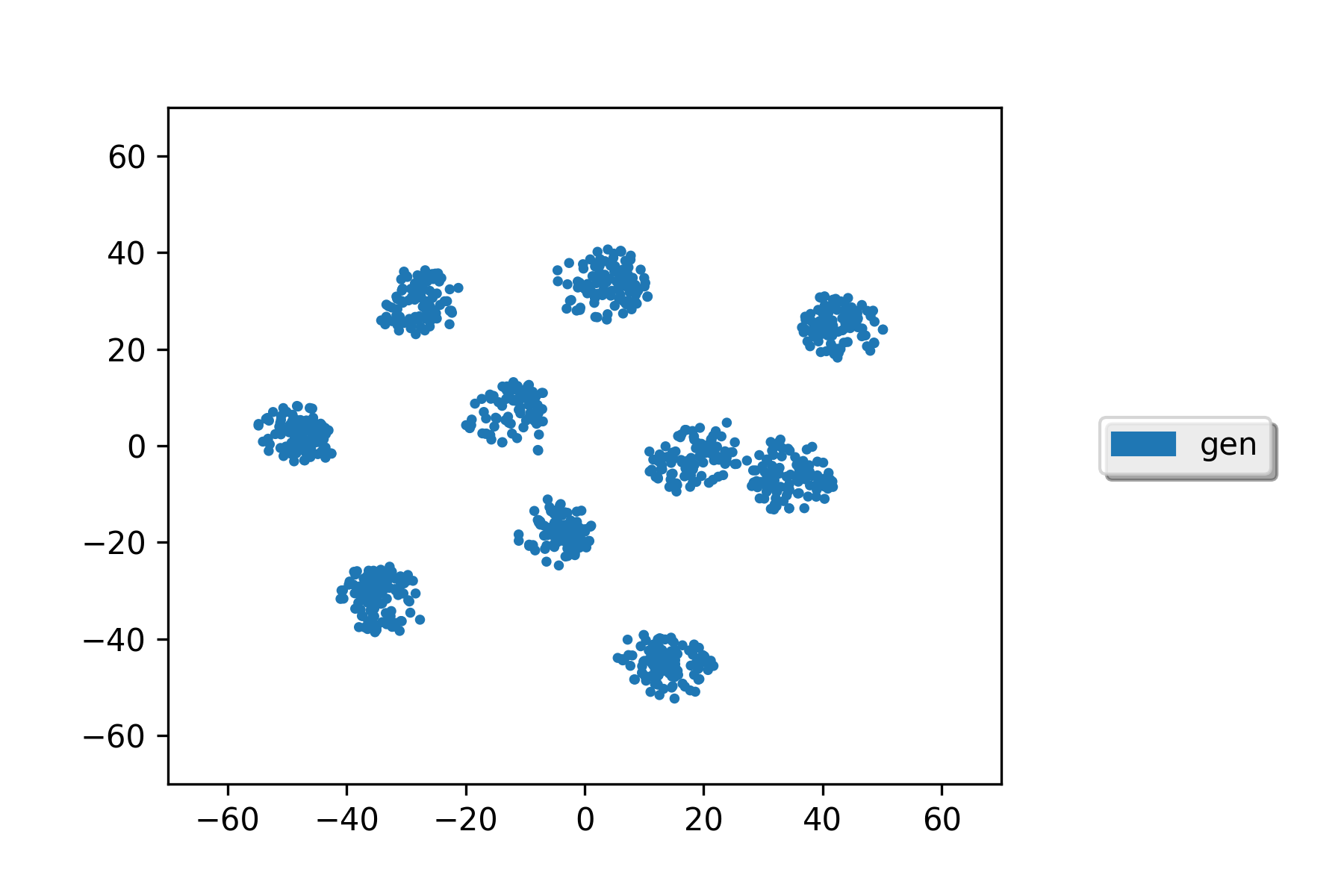}
        }
        \subfloat[Code generator (CIFAR-10)]{ 
                \includegraphics[width=0.33\linewidth]{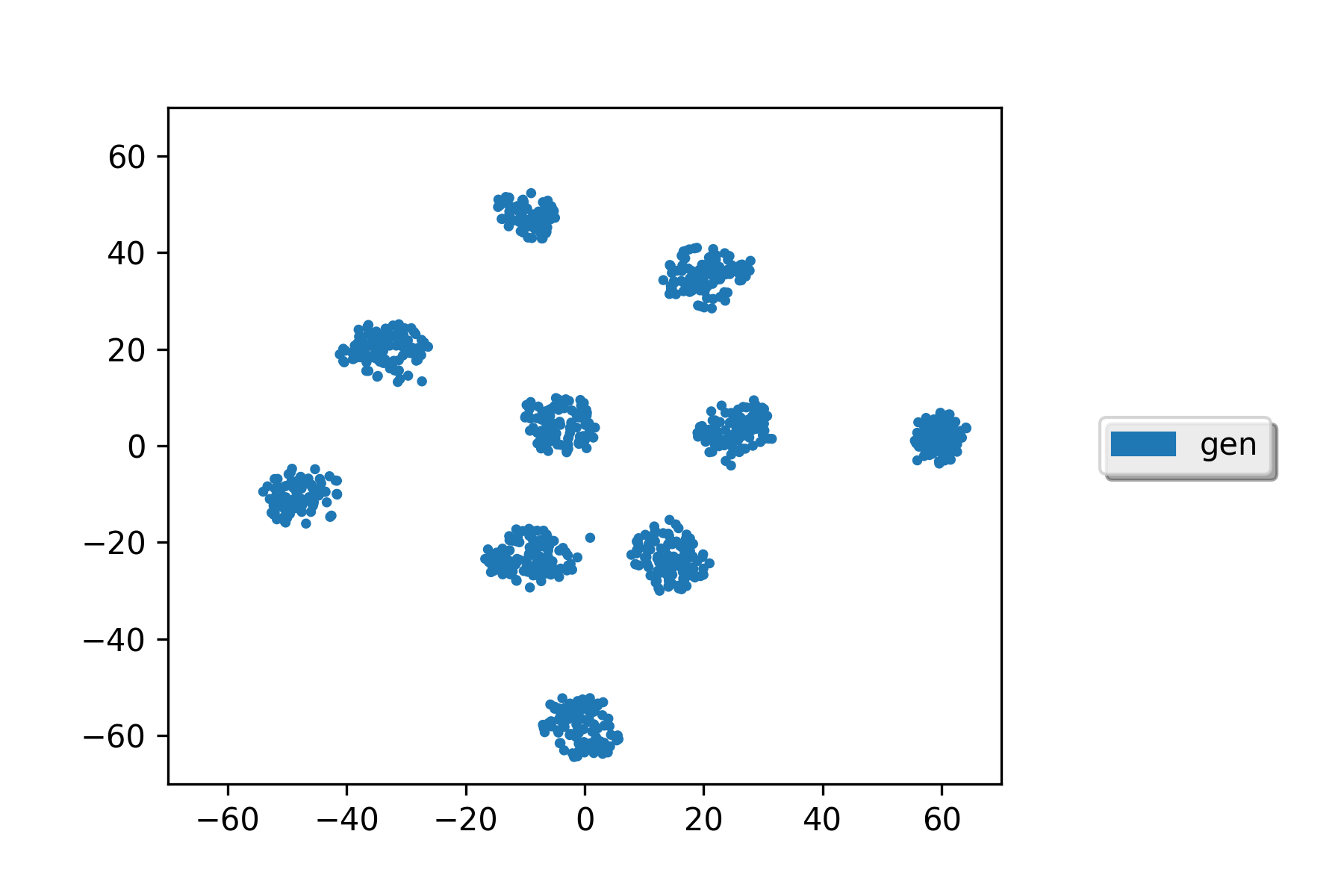}
        }
\caption{Visualization of the encoder output versus the code generator output
in the unsupervised setting.}
\label{fig:unsupervised_code_visualization}
\end{center}
\end{figure*}

Fig.~\ref{fig:unsupervised_code_visualization} further visualizes the latent code distributions at the output of the code generator and the encoder. Several observations can be made. First, the encoder is regularized well to produce an aggregated posterior distribution similar to that at the code generator output. Second, the code generator learns distinct conditional distributions according to the categorical label input $s$. Third, quite by accident, the encoder learns to cluster images of the same digit on MNIST, as has been confirmed in Fig.~\ref{fig:unsupervised_disentangled_results}.
As expected, such semantic clustering in code space is not obvious on more complicated SVHN and CIFAR-10, as is evident from the somewhat random assignment of latent codes to images of the same class or label.      

\vspace{-5pt}\section{Application: Text-to-Image Synthesis}
\label{sec:text2img}
This section presents an application of our model to text-to-image synthesis. We show that the code generator can transform the embedding of a sentence into a prior suitable for synthesizing images that match closely the sentence's semantics. To this end, we learn supervisedly the correspondence between images and their descriptive sentences using the architecture in Fig. \ref{fig:supervised_models}, where given an image-sentence pair, the sentence's embedding (which is a 200-D vector) generated by a pre-trained recurrent neural network is input to the code generator and the discriminator in image space as if it were the label information, while the image representation is learned through the autoencoder and regularized by the output of the code generator. As before, a 100-D Gaussian is placed at the input of the code generator to explain the variability of images given the sentence.   

\begin{figure}[tbp] 
    \centering
    \subfloat[This vibrant flower features lush red petals and a similar colored pistil and stamen]{%
        \label{fig:text2img_fixed_sentence_red}
        \includegraphics[width=0.45\linewidth]{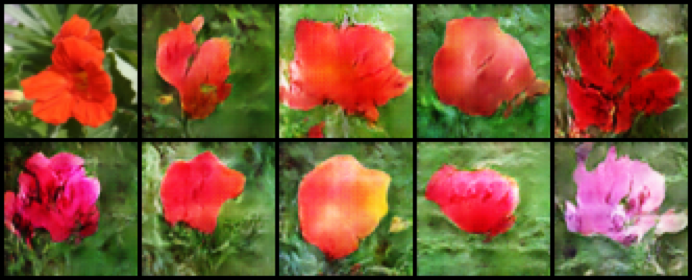}%
        }%
    \hspace{0.05\linewidth} 
    \subfloat[This flower has white and crumpled petals with yellow stamen]{%
        \label{fig:text2img_fixed_sentence_white}
        \includegraphics[width=0.45\linewidth]{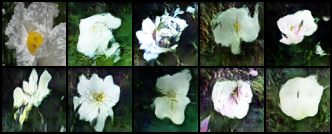}%
        }%
    \caption{Generated images from text descriptions.}
    \label{fig:text2img_fixed_sentence}
\end{figure}

\begin{figure}[tbp] 
    \centering
        \includegraphics[width=1.0\linewidth]{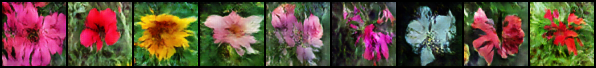}
    \caption{Generated images in accordance with the varying color attribute in the text description "The flower is pink in color and has petals that are rounded in shape and ruffled." From left to right, the color attribute is set to pink, red, yellow, orange, purple, blue, white, green, and black, respectively. Note that there is no green or black flower in the dataset.}
    \label{fig:text2img_replaced_sentence}
\end{figure}

The results in Fig. \ref{fig:text2img_fixed_sentence} present images generated by our model when trained on 102 Category Flower dataset \citep{Nilsback08}. The generation process is much the same as that described in Section \ref{sssec:supervised_exp}. It is seen that most images match reasonably the text descriptions. In Fig. \ref{fig:text2img_replaced_sentence}, we further explore how the generated images change with the variation of the color attribute in the text description. We see that most images agree with the text descriptions to a large degree. 

\section{Conclusion}
\label{sec:conclusion}

In this paper, we propose to learn a proper prior from data for AAE. Built on the foundation of AAE, we introduce a code generator to transform the manually selected simple prior into one that can better fit the data distribution. We develop a training process that allows to learn both the autoencoder and the code generator simultaneously. We demonstrate its superior performance over AAE in image generation and learning disentangled representations in supervised and unsupervised settings. We also show its ability to do cross-domain translation. Mode collapse and training instability are two major issues to be further investigated in future work.

\bibliographystyle{IEEEtran}
\bibliography{apsipa2018_conference}

\begin{thebibliography}{10}
\providecommand{\url}[1]{#1}
\csname url@samestyle\endcsname
\providecommand{\newblock}{\relax}
\providecommand{\bibinfo}[2]{#2}
\providecommand{\BIBentrySTDinterwordspacing}{\spaceskip=0pt\relax}
\providecommand{\BIBentryALTinterwordstretchfactor}{4}
\providecommand{\BIBentryALTinterwordspacing}{\spaceskip=\fontdimen2\font plus
\BIBentryALTinterwordstretchfactor\fontdimen3\font minus
  \fontdimen4\font\relax}
\providecommand{\BIBforeignlanguage}[2]{{%
\expandafter\ifx\csname l@#1\endcsname\relax
\typeout{** WARNING: IEEEtran.bst: No hyphenation pattern has been}%
\typeout{** loaded for the language `#1'. Using the pattern for}%
\typeout{** the default language instead.}%
\else
\language=\csname l@#1\endcsname
\fi
#2}}
\providecommand{\BIBdecl}{\relax}
\BIBdecl

\bibitem{larsen2015autoencoding}
A.~B.~L. Larsen, S.~K. Sønderby, H.~Larochelle, and O.~Winther, ``Autoencoding
  beyond pixels using a learned similarity metric,'' in \emph{Proceedings of
  the International Conference on Machine Learning (ICML)}, 2016, pp.
  1558--1566.

\bibitem{dilokthanakul2016deep}
N.~Dilokthanakul, P.~A. Mediano, M.~Garnelo, M.~C. Lee, H.~Salimbeni,
  K.~Arulkumaran, and M.~Shanahan, ``Deep unsupervised clustering with gaussian
  mixture variational autoencoders,'' \emph{arXiv preprint arXiv:1611.02648},
  2016.

\bibitem{makhzani2015adversarial}
A.~Makhzani, J.~Shlens, N.~Jaitly, I.~Goodfellow, and B.~Frey, ``Adversarial
  autoencoders,'' \emph{arXiv preprint arXiv:1511.05644}, 2015.

\bibitem{wu2016learning}
J.~Wu, C.~Zhang, T.~Xue, B.~Freeman, and J.~Tenenbaum, ``Learning a
  probabilistic latent space of object shapes via 3d generative-adversarial
  modeling,'' in \emph{Advances in Neural Information Processing Systems
  (NIPS)}, 2016, pp. 82--90.

\bibitem{kingma2013auto}
D.~P. Kingma and M.~Welling, ``Auto-encoding variational bayes,'' in
  \emph{Proceedings of the International Conference on Learning Representations
  (ICLR)}, 2013.

\bibitem{burda2015importance}
Y.~Burda, R.~Grosse, and R.~Salakhutdinov, ``Importance weighted
  autoencoders,'' \emph{arXiv preprint arXiv:1509.00519}, 2015.

\bibitem{hoffman2016elbo}
M.~D. Hoffman and M.~J. Johnson, ``Elbo surgery: yet another way to carve up
  the variational evidence lower bound,'' in \emph{Workshop in Advances in
  Approximate Bayesian Inference, NIPS}, 2016.

\bibitem{goyal2017nonparametric}
P.~Goyal, Z.~Hu, X.~Liang, C.~Wang, and E.~Xing, ``Nonparametric variational
  auto-encoders for hierarchical representation learning,'' \emph{arXiv
  preprint arXiv:1703.07027}, 2017.

\bibitem{tomczak2017vae}
J.~M. Tomczak and M.~Welling, ``Vae with a vampprior,'' \emph{arXiv preprint
  arXiv:1705.07120}, 2017.

\bibitem{chen2016infogan}
X.~Chen, Y.~Duan, R.~Houthooft, J.~Schulman, I.~Sutskever, and P.~Abbeel,
  ``Infogan: Interpretable representation learning by information maximizing
  generative adversarial nets,'' in \emph{Advances in Neural Information
  Processing Systems (NIPS)}, 2016, pp. 2172--2180.

\bibitem{johnson2016perceptual}
J.~Johnson, A.~Alahi, and L.~Fei-Fei, ``Perceptual losses for real-time style
  transfer and super-resolution,'' in \emph{Proceedings of the European
  Conference on Computer Vision (ECCV)}, 2016, pp. 694--711.

\bibitem{goodfellow2014generative}
I.~Goodfellow, J.~Pouget-Abadie, M.~Mirza, B.~Xu, D.~Warde-Farley, S.~Ozair,
  A.~Courville, and Y.~Bengio, ``Generative adversarial nets,'' in
  \emph{Advances in Neural Information Processing Systems (NIPS)}, 2014, pp.
  2672--2680.

\bibitem{berthelot2017began}
D.~Berthelot, T.~Schumm, and L.~Metz, ``Began: Boundary equilibrium generative
  adversarial networks,'' \emph{arXiv preprint arXiv:1703.10717}, 2017.

\bibitem{radford2015unsupervised}
A.~Radford, L.~Metz, and S.~Chintala, ``Unsupervised representation learning
  with deep convolutional generative adversarial networks,'' \emph{arXiv
  preprint arXiv:1511.06434}, 2015.

\bibitem{mao2017least}
X.~Mao, Q.~Li, H.~Xie, R.~Y. Lau, Z.~Wang, and S.~Paul~Smolley, ``Least squares
  generative adversarial networks,'' in \emph{Proceedings of the IEEE
  International Conference on Computer Vision (ICCV)}, 2017, pp. 2794--2802.

\bibitem{gulrajani2017improved}
I.~Gulrajani, F.~Ahmed, M.~Arjovsky, V.~Dumoulin, and A.~C. Courville,
  ``Improved training of wasserstein gans,'' in \emph{Advances in Neural
  Information Processing Systems (NIPS)}, 2017, pp. 5767--5777.

\bibitem{salimans2016improved}
T.~Salimans, I.~Goodfellow, W.~Zaremba, V.~Cheung, A.~Radford, and X.~Chen,
  ``Improved techniques for training gans,'' in \emph{Advances in Neural
  Information Processing Systems (NIPS)}, 2016, pp. 2234--2242.

\bibitem{Nilsback08}
M.-E. Nilsback and A.~Zisserman, ``Automated flower classification over a large
  number of classes,'' in \emph{Proceedings of the Indian Conference on
  Computer Vision, Graphics and Image Processing}, 2008.

\end{thebibliography}

\vskip2pc

\noindent \large \textbf{Biographies} (no more than 150 words each)

\vskip2pc

\noindent\normalsize\textbf{Hui-Po Wang} received his B.S. degree in computer science from National Tsing Hua University in 2017. He is currently pursuing his M.S. degree in National Chiao Tung University. His research interest lies in deep generative models, unsupervised domain adaptation, and related vision applications.

\vskip2pc

\noindent\textbf{Wen-Hsiao Peng} received his Ph.D. degree from National Chiao Tung University (NCTU), Taiwan in 2005. He was with Intel Microprocessor Research Laboratory, USA from 2000 to 2001. Since 2003, he has actively participated in the ISO/IEC and ITU-T video coding standardization process. He is a Professor with the Computer Science Department, NCTU, and was a Visiting Scholar with the IBM Thomas J. Watson Research Center, USA, from 2015 to 2016. His research interests include video/image coding and multimedia analytics. Dr. Peng is an IEEE CASS VSPCTC member. He was Publication Chair for 2019 IEEE ICIP, Technical Program Co-chair for 2011 IEEE VCIP, 2017 IEEE ISPACS and 2018 APSIPA ASC, Area Chair for IEEE ICME and VCIP, Track Chair for IEEE ISCAS. He served as Guest Editor/SEB Member for IEEE JETCAS and Associate Editor for IEEE TCSVT. He was elected Chair-elect of IEEE CASS VSPCTC and Distinguished Lecturer of APSIPA.

\vskip2pc

\noindent\textbf{Wei-Jan Ko} received the B.S. degree in traffic science from Central Police University and the M.S. degree in computer science from National Chiao Tung University (NCTU), Taiwan, in 2014 and 2018, respectively. He is currently a doctoral student in computer science at NCTU. He has been a teaching assistant in Introduction of Artificial intelligence course in NCTU, and instructor of Artificial intelligence and Deep Learning in NCTU MOOC(Massive open online course) platform. His current research interests include Deep Generative Model and Intelligent Systems with smart city.

\vskip2pc

\noindent \large \textbf{LIST OF FIGURES AND TABLES (for authors submitting their work in Word)}

\vskip1pc

\noindent \normalsize Fig. 1. Wide-band push--pull amplifier scheme.

\noindent{Table 1. Measured performances in different operating conditions.}

\end{document}